\documentclass[11pt]{article}

\usepackage[]{acl}
\usepackage{fancyvrb}
 \usepackage{subcaption}
\usepackage{times}
\usepackage{latexsym}
\usepackage{graphicx}
\usepackage{listings}
\usepackage{booktabs}
\usepackage{longtable}
\usepackage{multirow}
\usepackage{colortbl}    % For \cellcolor
\usepackage[table]{xcolor} % Better color support
\usepackage{float}
\usepackage{stfloats}
\usepackage{chngcntr}
\usepackage{tabularx}
\usepackage{threeparttable}
\usepackage{booktabs}
\usepackage[a4paper, margin=2.5cm]{geometry}
\usepackage{caption}
\usepackage{amssymb}
\usepackage{placeins}
\usepackage{stfloats}
\usepackage{xcolor}
\definecolor{codegray}{gray}{0.95}
\usepackage{chngcntr}
\counterwithout{table}{section}
\counterwithout{figure}{section}

\usepackage[T1]{fontenc}
\usepackage[utf8]{inputenc}
\usepackage{microtype}
\usepackage{inconsolata}
\usepackage{graphicx}
\title{Evaluating LLM-Generated Versus Human-Authored Responses in Role-Play Dialogues}

\author{
  Dongxu Lu, Johan Jeuring, Albert Gatt \\
  Utrecht University, Utrecht, The Netherlands \\
  \texttt{\{d.lu, j.t.jeuring, a.gatt\}@uu.nl}
}

\begin{document}
\maketitle
\begin{abstract}
Evaluating large language models (LLMs) in long-form, knowledge-grounded role-play dialogues remains challenging. This study compares LLM-generated and human-authored responses in multi-turn professional training simulations through human evaluation ($N=38$) and automated \textit{LLM-as-a-judge} assessment. Human evaluation revealed significant degradation in LLM-generated response quality across turns, particularly in naturalness, context maintenance and overall quality, while human-authored responses progressively improved. In line with this finding, participants also indicated a consistent preference for human-authored dialogue. These human judgements were validated by our automated \textit{LLM-as-a-judge} evaluation, where \textsc{Gemini 2.0 Flash} achieved strong alignment with human evaluators on both zero-shot pairwise preference and stochastic 6-shot construct ratings, confirming the widening quality gap between LLM and human responses over time. Our work contributes a multi-turn benchmark exposing LLM degradation in knowledge-grounded role-play dialogues and provides a validated hybrid evaluation framework to guide the reliable integration of LLMs in training simulations.

\end{abstract}

\section{Introduction}
The rapid advancement of large language models (LLMs) has led to their increasing application in role-play dialogue systems across diverse domains, such as healthcare \cite{kenny_virtual_2024} and education \cite{an_developing_2024}. A key application of these systems is professional skills training in real-life scenarios. For instance, \citet{kenny_virtual_2024} integrated an LLM to role-play as a virtual patient, generating responses with realistic medical symptoms for clinical training. Similarly, in education, LLM-based pedagogical agents enhance students' collaborative learning \cite{an_developing_2024}. Such training simulations require that LLMs not only generate contextually appropriate responses but also remain grounded in given character profiles and domain-specific knowledge. Furthermore, these interactions are often guided by pedagogical goals for an immersive and effective training experience~\cite{gousseva_creating_2024}.

However, despite the growing deployment of LLMs in these applications, their evaluation remains a significant challenge \cite{chen_oscars_2024}. A critical gap exists in the availability of high-quality open-source datasets and benchmarks designed for knowledge-grounded role-play dialogue settings \cite{wang_rolellm_2024}, which are essential for guiding simulation optimisation \cite{wu_raiden_2025}. Most available benchmarks focus either on open-domain conversations, which prioritise generating engaging, open-ended dialogue \cite{feng_infusing_2024}, or on task-oriented conversations, where an agent assists a user with a concrete task, such as booking a hotel \cite{mo_hiertod_2025}. Neither paradigm fully addresses the demands of our targeted dialogue systems.

Moreover, the inadequacy of current evaluation methods is further intensified by their tendency to disregard the multi-turn nature of dialogues. The prevailing benchmark, MT-Bench \cite{zheng_judging_2023}, predominantly evaluates LLMs on a coarse-grained level using minimal turn-taking, such as two-turn dialogues. Such a methodology fails to account for performance degradation over longer interactions, a phenomenon demonstrated by \citet{liu_lost_2024}, who found that even advanced LLMs perform significantly worse in multi-turn contexts. Accordingly, a comprehensive multi-turn evaluation is essential to gain an understanding of how LLMs function in long-form dialogue systems.

In this paper, we aim to address these gaps by conducting a multi-turn comparative analysis of LLM-generated responses and human-authored responses within knowledge-grounded training simulations. By systematically analysing the quality of LLM-generated responses throughout these interactions, we aim to determine if LLMs can capture the nuances of real-life conversations and maintain comparable performance to humans over time. This motivates our research question:

\textit{RQ: How do LLM-generated and human-au\-thored responses compare in knowledge-grounded role-play conversations over multiple turns?} 

To answer this question, we conducted two experiments, using pre-existing, human-authored conversations as a baseline and prompting a fine-tuned LLM to generate an alternative for each turn. Experiment 1 (Section \ref{sec:exp1}) focused on a detailed human evaluation of a single scenario, analysing constructs adapted from the USR framework \cite{mehri_usr_2020} and capturing expert opinions through a focus group. Building on these initial results, Experiment 2 (Section \ref{subsec:exp2}) tested the generalisability of our findings across multiple diverse scenarios by applying automated evaluation using \textit{LLM-as-a-judge}.

Across both experiments, our results revealed a significant degradation in the perceived quality of LLM-generated responses as the dialogue progressed, while human-authored responses were consistently perceived as higher quality. These findings provide critical insights for the future design of knowledge-grounded training simulations and effective integration of LLMs into role-play dialogue systems.

\section{Related Work}

\subsection{Automatic Dialogue Evaluation}
The evolution of dialogue evaluation methods reflects a growing recognition of conversation complexity. Traditional metrics, such as BLEU \cite{papineni_bleu_2002}, ROUGE \cite{lin_rouge_2004}, and METEOR \cite{banerjee_meteor_2005}, established foundational approaches measuring lexical overlap. However, their reliance on n-gram matching proved insufficient for capturing dialogue-specific qualities, particularly in open-domain interactions where a wide variety of valid responses is possible \cite{liu_how_2016}. Consequently, these metrics often show a limited correlation with human judgments of conversational quality.

To overcome these limitations, automated evaluation metrics emerged that applied learned representations of utterances to assess dialogues. These methods moved beyond surface-level text matching to analyse semantic coherence and quality. For instance, some approaches use large pre-trained language models to generate latent representations of utterances and then train classifiers on these representations for evaluation \cite{mehri_usr_2020, sinha_learning_2020}. Among these, the USR framework \cite{mehri_usr_2020} is particularly relevant to our work. Its emphasis on dialogue contexts and facts, validated through evaluations on knowledge-grounded datasets such as PersonaChat \cite{zhang_personalizing_2018} and Topical-Chat \cite{gopalakrishnan_topical-chat_2019}, makes it applicable for use cases like professional skill training.

\subsection{LLM-as-a-Judge Paradigm}
\textit{LLM-as-a-judge} is a promising paradigm to simulate the depth and granularity of human evaluation \cite{zheng_judging_2023}. This approach typically prompts an LLM to perform either point-wise scoring or pairwise comparisons \cite{li_generation_2025}. Comparative analyses have shown that pairwise assessment consistently outperforms point-wise scoring in aligning with human annotations \cite{kim_prometheus_2024}. For instance, PairEval \cite{park_paireval_2024} demonstrated that moderately-sized language models can achieve human-level agreement in pairwise response comparisons. 

Despite this promise, a known challenge is that LLM judges can exhibit biases, such as a preference for their own style of generation \cite{panickssery_llm_2025}. To address this, recent work has focused on developing specialised judge models, such as Prometheus-2 \cite{kim_prometheus_2024} and JudgeLM \cite{zhu_judgelm_2025}, designed to improve objectivity.

\subsection{Multi-Turn Dialogue Evaluation}
While the \textit{LLM-as-a-judge} paradigm offers an automated assessment alternative, evaluating conversations that span over multiple turns introduces challenges related to context retention and interaction dynamics. MT-Bench \cite{bai_mt-bench-101_2024} pioneered by employing LLM judges to assess the multi-turn capabilities of LLMs in open-domain settings, such as perceptivity, adaptability, and interactivity. For retrieval-augmented dialogues, CORAL \cite{cheng_coral_2025} measured citation accuracy during topic transitions over multiple turns. 

Furthermore, game-based benchmarks have emerged to assess the multi-turn capabilities and interaction dynamics of LLMs in goal-oriented gameplay settings. For instance, TextArena \cite{guertler_textarena_2025} places agents in competitive scenarios, using game outcomes as a direct measure of capability, while benchmarks like Clembench \cite{chalamalasetti_clembench_2023} and GameBench \cite{costarelli_gamebench_2024} investigate collaborative and strategic reasoning skills through interactive gameplay.

A limitation across these benchmarks is their handling of long-range context information. As conversations exceed the typical context windows, they can suffer the "lost in the middle" phenomenon \cite{liu_lost_2024}, leading to significant degradation in evaluation fidelity \cite{hankache_evaluating_2025}. Notably, benchmarks such as MultiChallenge \cite{deshpande_multichallenge_2025} have evaluated in-contextual reasoning, demonstrating that even state-of-the-art models struggle with complex sequential contexts.

\section{Experiment 1: Human Evaluation}
\label{sec:exp1}
The main goal of this experiment was to address our main research question, the comparative quality of LLM-generated and human-authored responses in a multi-turn, knowledge-grounded role-play. To guide our investigation, we formulated two sub-research questions (Sub-RQs):

\begin{description}
    \item[SubRQ1:] \textit{How do the quality perceptions of LLM-generated and human-authored responses change as a dialogue progresses over turns?}
    \item[SubRQ2:] \textit{What key factors most strongly influence participants’ perceptions of response quality?}
\end{description}

\subsection{Participants}
We recruited 38 participants: 19 via convenience sampling through the researchers' social networks, and the remaining 19 from the Prolific platform. All participants had to be fluent in English. Participants were aged 22 to 55 years, and 25 identified as male and 14 as female. 39.5\% ($n=15$) held an undergraduate degree, 57.9\% ($n=22$) held a Master's degree, and 2.6\% ($n=1$) held a PhD.

\subsection{Materials and Stimuli}
\label{subsubsec:exp1_scenario}

\begin{figure}[h!]
    \centering
    \includegraphics[width=0.98\linewidth]{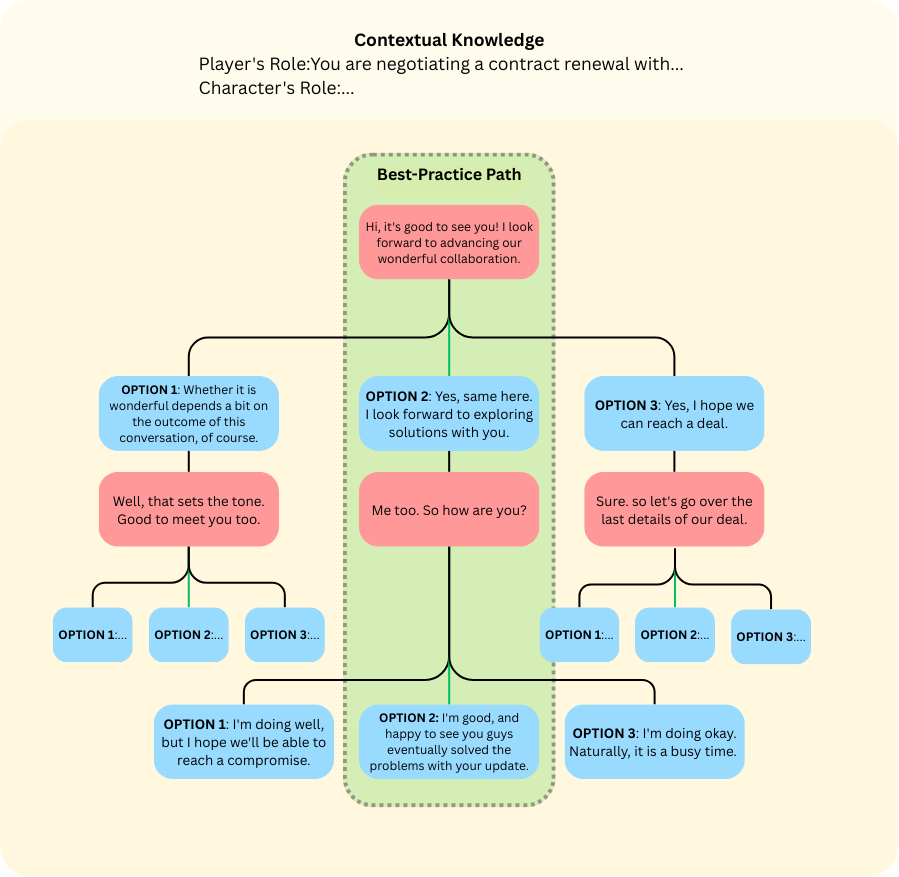}
    \caption{Overview of the training scenario structure. Each scenario includes contextual knowledge (e.g., player and character roles) and a dialogue tree. Agent statements are represented by red nodes and player choices by blue nodes. The highlighted sequence represents the `best-practice path'.}
    \label{fig:simulation}
\end{figure}

To generate structured and comparable conversations, we collaborated with a company specialising in communication training software. Their platform provides conversational training simulations built around various scenarios that place a user in a professional situation requiring a specific conversational skill. An earlier version of this platform has been described as a serious game for communication skills \cite{jeuring_communicate_2015}. Each simulation is structured as a decision tree, where the user makes choices to navigate the dialogue with a virtual agent. Within this structure, an optimal sequence of choices forms a \textit{best-practice path} designed to achieve the desired conversational outcome as shown in Figure \ref{fig:simulation}.

For this study, we selected a single scenario focused on negotiation skills. The full simulation of this scenario consists of 24 conversational exchanges between the user and the agent. The agent's first turn, which serves as a greeting without any preceding dialogue, was excluded from our analysis. This resulted in a total of 23 exchanges for which we generated two distinct sets of agent responses to serve as our stimuli:

\begin{itemize}
\item \textbf{Human-Authored Responses:} The agent's pre-scripted utterances from the \textit{best-practice path} were used as the benchmark for high-quality, human-authored content.

\item \textbf{LLM-Generated Responses:} To create a parallel dialogue, we prompted a fine-tuned \textsc{Llama 3} model (see Table~\ref{tab:model_comparison} for model specifications and Table~\ref{tab:hyperparameters} for fine-tuning details) to generate an agent response at each turn along the same \textit{best-practice path}. The model was fine-tuned on a pre-selected set of 90 high-quality conversation scenarios to ensure relevant output in the desired style. 

The prompt provided the model with the scenario's contextual knowledge as well as the dialogue context comprising the preceding conversation history and the next statement in the "best-practice path". This process ensured that the LLM generated a complete, parallel dialogue that followed the same sequence of turns as the human-authored version. The prompt template is given in Figure~\ref{fig:prompt_generation}.
\end{itemize}

Table~\ref{tab:dialogue_example} presents a side-by-side comparison of two response types for 23 agent turns in the dialogue. Pilot testing indicated that evaluating 23 response pairs in a single session could induce cognitive fatigue. To mitigate this, we partitioned the turns into Session A (12 turns) and Session B (11 turns). Each participant was randomly assigned to evaluate responses in only one of the two sessions.

\subsection{Measures}
Participants rated each agent's response on the following six quality dimensions adapted from the USR metric framework \cite{mehri_usr_2020}. 

\begin{description}
    \item[Understandable (0--1)]"Is the response understandable in the context of the history?"
    \item[Natural (1--3)] "Is the response naturally written?"
    \item[Maintains Context (1--3)] "Does the response serve as a valid continuation of the conversation history?"
    \item[Interesting (1--3)] "Is the response dull/interesting?" 
    \item[Uses Knowledge (0--1)]  "Given the contextual knowledge of the scenario (e.g. character role's description, scenario background) that the response is conditioned on, how well does the response use the knowledge?" (Adapted from \citet{mehri_usr_2020} to fit our context.) 
    \item[Overall Quality (1--5)] "Given your answers above, what is your overall impression of this utterance?" 
\end{description}

\subsection{Procedure}

After providing informed consent, participants received instructions on the study's background and their tasks. Each participant was randomly assigned to either session A or session B. The experiment consisted of two main tasks for each participant.

\paragraph{In-simulation pairwise preference} At each agent's turn in the dialogue simulation, participants were presented with two responses: one LLM-generated and one human-authored. To mitigate order effects, the presentation order of these two responses was randomised for each turn. They were then asked, "Which response do you think fits best within the conversation?" and indicated their preference. This process, as shown in Figure \ref{fig:pre_procedure}, was repeated for all turns in their assigned session.

\paragraph{Post-simulation rating} After completing the simulation, participants were directed to a questionnaire where they were shown the same response pairs in the simulation context again and asked to rate each response on chosen constructs.

The experiment concluded after these two tasks were completed.

\subsubsection{Focus Group}
\label{subsubsec:exp1_focus_group}
Following the main experiment, we conducted a focus group with two instructional designers from the company to gather expert qualitative insights. During the session, designers were shown five dialogue exchanges containing both LLM-generated and human-authored responses. We performed a guided discussion to understand which design aspects they value most, and to identify important qualities potentially missed by our quantitative constructs.

\subsection{Data Analysis}
We analysed the data from Experiment 1 using both quantitative and qualitative methods to address our sub-research questions.

To answer \textbf{SubRQ1}, we fitted linear mixed-effects models (LMMs) with response \texttt{Condition} (LLM-generated vs. human-authored), conversational \texttt{Turn}, and their interaction as fixed effects, including random intercepts for individual participants assigned to session A and session B. This allowed for a multi-turn analysis on whether the relative quality of response conditions changed as the dialogue progressed. To examine trends in perceived response quality, we plotted the mean rating per turn for each response condition and overlaid a corresponding trend line generated by a simple linear model (Ordinary Least Squares, OLS).

We analysed the proportion of participants who expressed a preference for the LLM's response at each of the 23 conversational turns, identifying representative turns for each response condition.

To address \textbf{SubRQ2}, we conducted a mixed-methods analysis. First, a Spearman correlation analysis was performed among the evaluated constructs to identify which dimensions were most strongly associated with \textit{Overall Quality} ratings. Second, the focus group session was conducted via Google Meet and automatically transcribed using its built-in functionality. The resulting transcript was then corrected for errors by the first author to ensure accuracy. We then analysed this transcript using thematic analysis. This process involved two stages. First, we inductively coded the transcript by labelling key statements and concepts. Second, we grouped and refined these codes into the resulting themes presented in our findings.

\subsection{Results}
\subsubsection{Multi-turn Analysis of Perceived Quality}
Our analysis of human-annotated ratings revealed how perceived quality changed over the course of a conversation. We found significant main effects for the conversational \texttt{Turn}, but the most critical finding was a consistent and statistically significant interaction between \texttt{Condition} and \texttt{Turn}. The key interaction effects are presented in Table \ref{tab:interaction_results}, while the full model statistics are available in Table \ref{tab:lmm_results}.

\begin{figure*}[h]
  \centering
  \begin{subfigure}[b]{0.32\linewidth}
    \includegraphics[width=\linewidth]{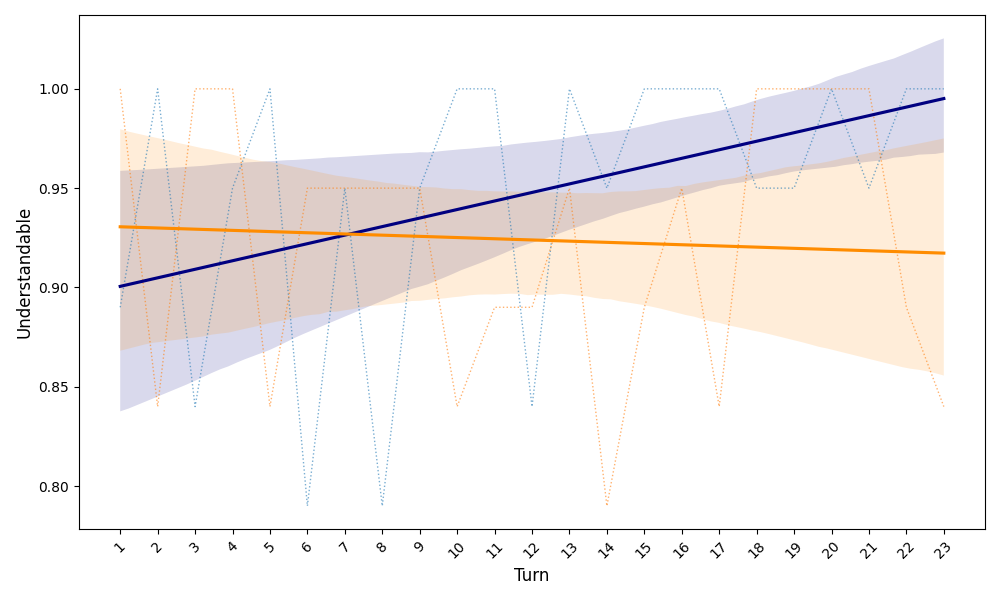}
    \caption{Understandable}
  \end{subfigure}
  \hfill
  \begin{subfigure}[b]{0.32\linewidth}
    \includegraphics[width=\linewidth]{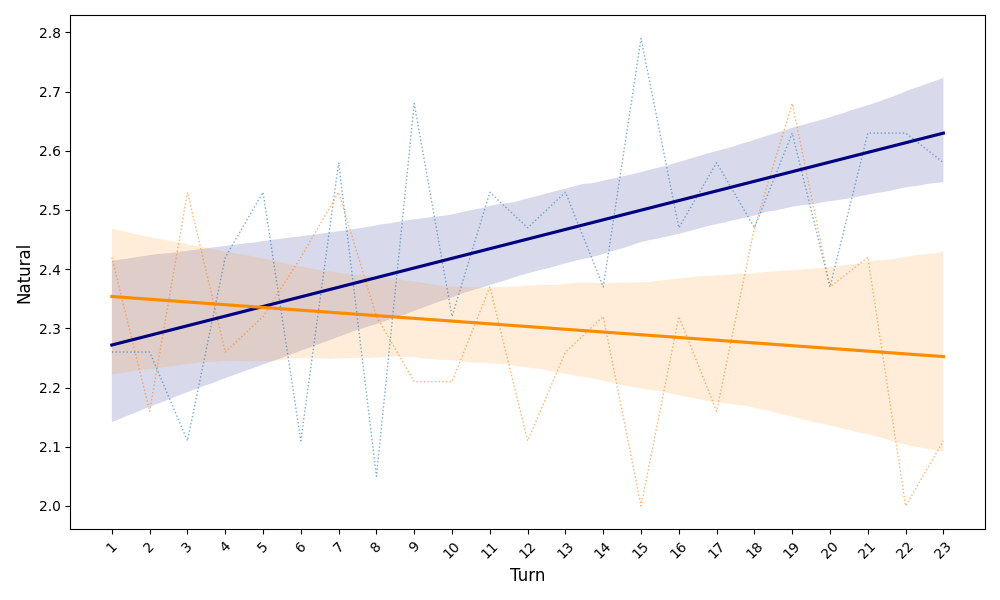}
    \caption{Natural}
  \end{subfigure}
  \hfill
  \begin{subfigure}[b]{0.32\linewidth}
    \includegraphics[width=\linewidth]{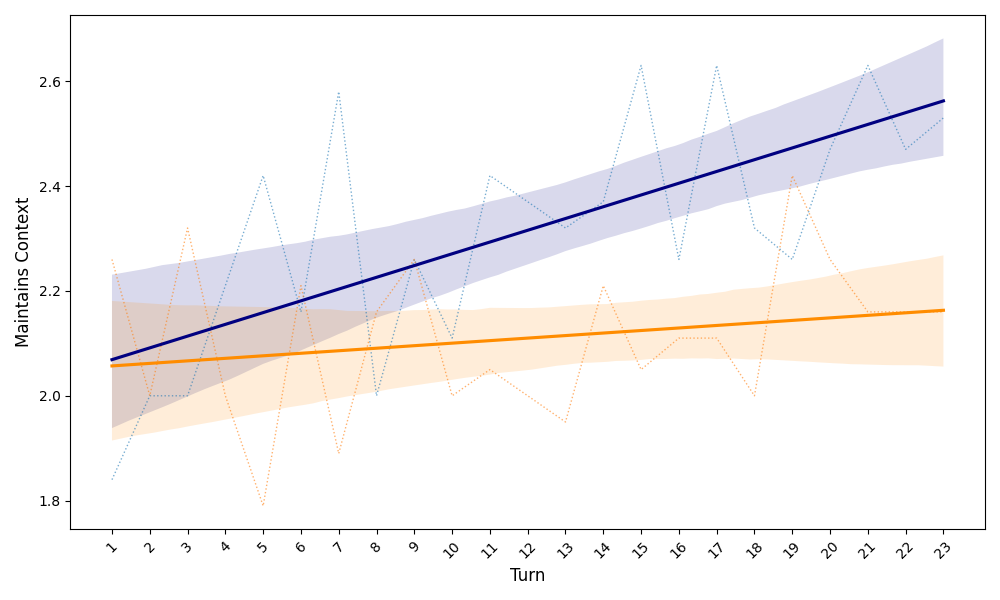}
    \caption{Maintains Context}
  \end{subfigure}
  
  \vspace{1em}
  
  \begin{subfigure}[b]{0.32\linewidth}
    \includegraphics[width=\linewidth]{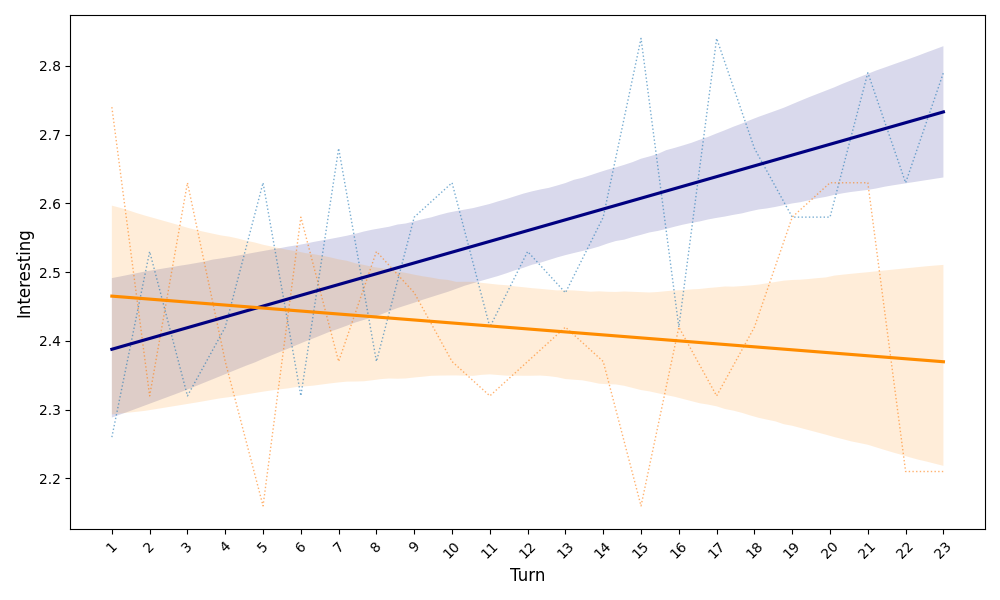}
    \caption{Interesting}
  \end{subfigure}
  \hfill
  \begin{subfigure}[b]{0.32\linewidth}
    \includegraphics[width=\linewidth]{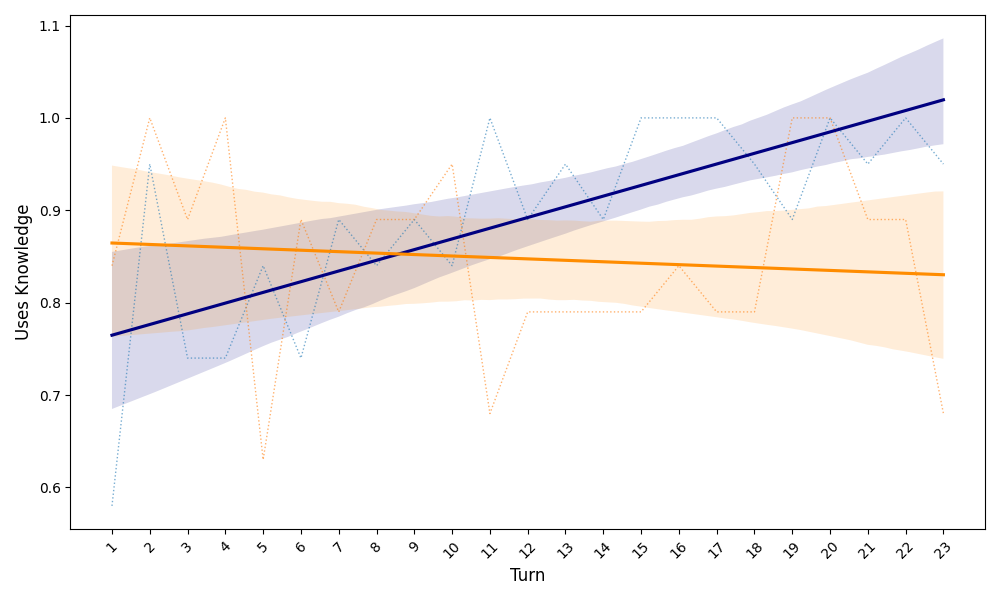}
    \caption{Uses Knowledge}
  \end{subfigure}
  \hfill
  \begin{subfigure}[b]{0.32\linewidth}
    \includegraphics[width=\linewidth]{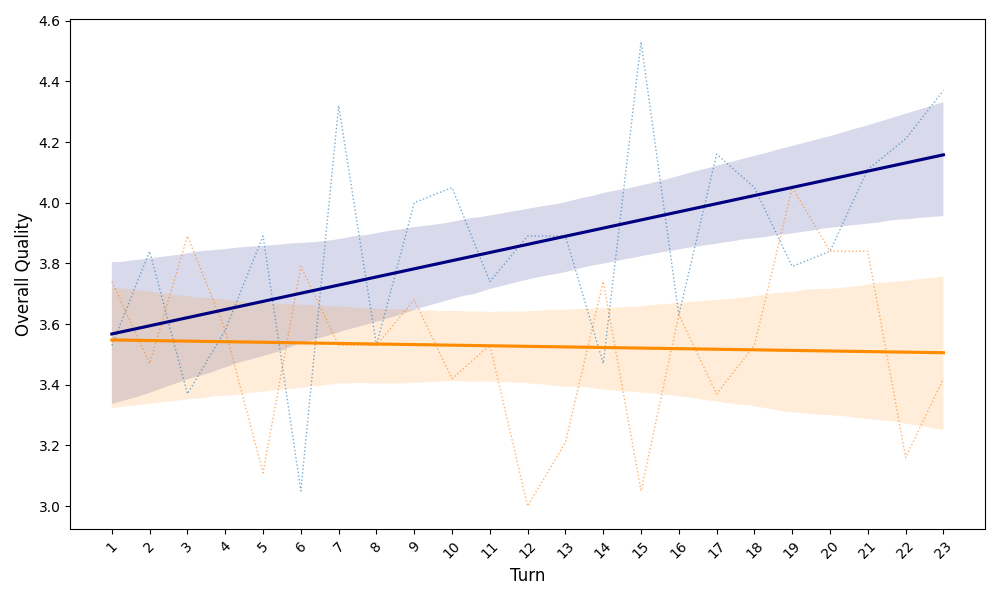}
    \caption{Overall Quality}
  \end{subfigure}

  \caption{Perceived quality ratings of LLM-generated (orange) and human-authored (blue) responses over turns. Dotted lines connect the mean rating at each turn for each response condition. Solid lines represent the overall linear regression (OLS) trend.}
  \label{fig:main_construct_plots}
\end{figure*}

The analysis first revealed a significant positive main effect of \texttt{Turn} across all six quality constructs. This indicates that, on average, responses were perceived more favourably as the dialogues progressed. However, this trend was qualified by the significant \texttt{Condition × Turn} interaction, which points to a crucial difference in performance trajectories. As shown in Figure \ref{fig:main_construct_plots}, while human-authored responses showed a slight but steady increase in perceived quality throughout the dialogue, the LLM-generated responses declined in quality over time. This degradation in LLM performance was particularly notable for key conversational qualities, including \textit{Overall Quality} ($\beta=-0.029,p=.001$), \textit{Natural} ($\beta = -0.021,p<.001$), and \textit{Maintains Context} ($\beta=-0.020,p < .001$).

The main effect for \texttt{Condition} revealed a more nuanced performance. Across five of the six constructs, including \textit{Overall Quality}, there was no statistically significant difference in the average ratings between the LLM-generated and human-authored response conditions. Although LLM showed superior performance in \textit{Uses Knowledge}, this initial advantage was progressively negated by accumulating conversational contexts, as proved by the significant negative interaction ($\beta=-0.013,p<.001$).

\begin{table*}[h]
\centering
\caption{Results of the LMMs for each quality construct. The table displays the fixed-effect coefficients ($\beta$), standard errors ($SE$), $z$-values, and $p$-values for the \texttt{Condition × Turn} interaction.}
\label{tab:interaction_results}
\begin{tabular}{llcccc}
\hline
\textbf{Construct} & \textbf{Effect} & \textbf{$\beta$} & \textbf{$SE$} & \textbf{z-value} & \textbf{p-value} \\ \hline
\textit{Understandable} & \texttt{Condition × Turn} & -0.005 & 0.002 & -2.022 & .043\\ 
\textit{Natural}        & \texttt{Condition × Turn} & -0.021 & 0.005 & -3.863 & <.001\\ 
\textit{Maintains Context} & \texttt{Condition × Turn} & -0.020 & 0.005 & -3.706 & <.001\\ 
\textit{Interesting}    & \texttt{Condition × Turn} & -0.018 & 0.006 & -2.930 & .003\\ 
\textit{Uses Knowledge} & \texttt{Condition × Turn} & -0.013 & 0.003 & -4.371 & <.001\\ 
\textit{Overall Quality} & \texttt{Condition × Turn} & -0.029 & 0.009 & -3.363 & .001\\ \hline
\end{tabular}
\end{table*}

\subsubsection{Participant Preference}
\begin{figure}[H]
    \centering
    \includegraphics[width=1.0\linewidth]{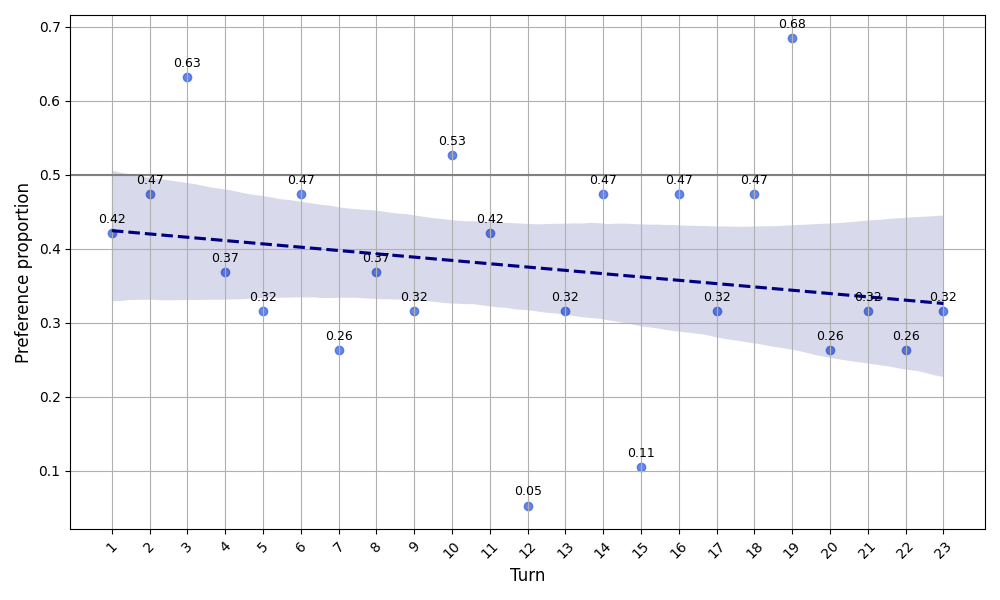}
    \caption{Proportion of participants preferring the LLM-generated response over turns. Data points represent the preference proportion at each turn. The dashed OLS trend line and its shaded 95\% confidence interval illustrate that there was no statistically significant trend over time ($p>.5$).}
    \label{fig:llm-preferences}
\end{figure}

The analysis of participants' preferences indicates a general preference for human-authored responses, with the LLM-generated response being selected by less than half of the participants in 20 out of the 23 turns. However, the preference for the LLM-generated responses varied considerably throughout the dialogue. In several instances, the LLM-generated response was strongly favoured, achieving a majority preference in three turns and reaching a peak of 68\% at turn 19. Conversely, its performance was viewed unfavourably in other turns, dropping to a preference of only 5\% at turn 12 and 11\% at turn 15. While a visual inspection of the data points suggests a slight downward trend for LLM preference over turns, as shown in Figure \ref{fig:llm-preferences}, the linear regression results suggest that this trend is not statistically significant ($\beta=-0.01, p = 0.504$).

\subsubsection{Correlation Analysis}
\label{subsec:correlation_results}

\begin{figure}[H]
    \centering
    \includegraphics[width=\linewidth]{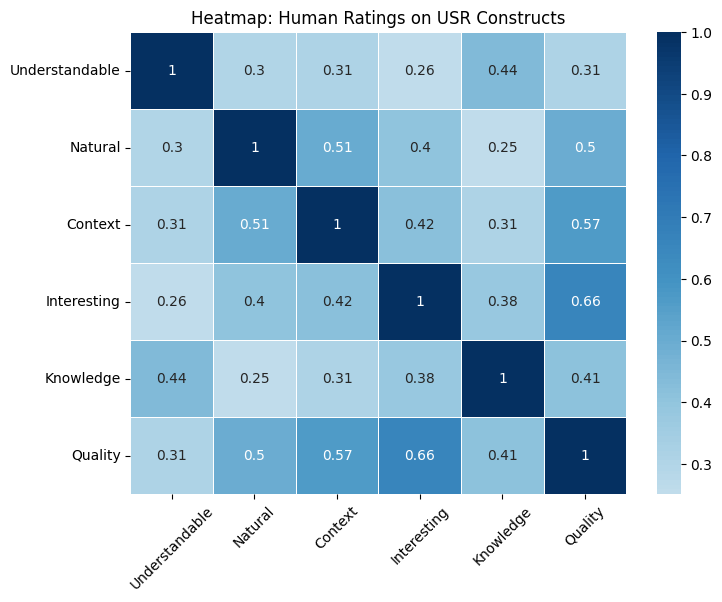}
    \caption{Heatmap of Spearman correlation coefficients ($r_s$) between all rated quality constructs.}
    \label{fig:correlation_heatmap}
\end{figure}

Spearman correlation analysis was used to examine the relationships between all rated constructs across individual ratings ($N=874$). The full correlation matrix is visualised in the heatmap in Figure~\ref{fig:correlation_heatmap}.

The analysis revealed that participants' ratings of \textit{Overall Quality} were most strongly and positively correlated with a response's perceived \textit{Interestingness} ($r_s = .66, p < .001$), followed by how well it \textit{Maintains Context} ($r_s = .57, p < .001$) and its \textit{Naturalness} ($r_s = .50, p < .001$).

\subsubsection{Qualitative findings of focus group}

The thematic analysis of transcribed focus group discussions revealed five themes regarding the essential attributes of effective dialogue responses in a training simulation context:

\paragraph{Natural Flow}
Both designers emphasised that responses with unnatural phrasing or awkward transitions disrupt user immersion and undermine the learning experience. 

\paragraph{Contextual Fit}
High-quality responses must reflect both dialogue history and scenario background. Even realistic-sounding replies were criticised if they contradicted established facts or ignored narrative context.

\paragraph{Tone Appropriateness}
Tone should align with the scenario’s emotional dynamics. Defensive or uncooperative tones were seen as misaligned, even if contextually plausible.

\paragraph{Pedagogical Nudging}
Responses should subtly guide users toward learning objectives without being overly directive.

\paragraph{Sentence Length}
Designers favoured concise replies with clear intent early in the sentence to support pacing and clarity.

\section{Experiment 2: Automated Evaluation}
\label{subsec:exp2}
Experiment 1 provided strong evidence that the quality of LLM-generated responses degraded over the course of a long-form dialogue and that human-authored responses were consistently preferred over LLM-generated ones. To investigate whether these findings generalise across broader conversational contexts, we designed a second experiment using an \textit{LLM-as-a-judge} for automated evaluation.

The experiment was structured in two phases. First, we conducted a validation phase, where we systematically evaluated how well LLMs mimic the human judgements from Experiment 1. Second, after identifying the evaluation method that reached the highest agreement with human judgements, we proceeded to the generalisation phase. In this phase, we applied the validated method to assess responses across three additional conversational scenarios, each designed for a different communication skill: motivational interviewing, selling, and consulting. Each selected scenario spans over 30 turns to ensure sufficient length for robust multi-turn analysis. This two-phase approach was implemented for two distinct evaluation tasks: construct rating and pairwise preference.

\subsection{Task 1: Construct Rating}
\label{subsubsec:exp2_validation}
\paragraph{Validation on Ground-Truth}
We started by determining a plausible setup for an LLM judge to predict the fine-grained quality ratings from Experiment 1. We prompted LLMs with varying capabilities, detailed in Table~\ref{tab:model_comparison} (\textsc{Llama 3.1 8B}, \textsc{Mistral 7B}, \textsc{Phi-3 Medium 14B}, and \textsc{Gemini 2.0 Flash}), to rate responses on the same six quality constructs in zero-shot settings. Each prompt provided the LLM with the complete context available to human annotators, including the scenario background, character roles, and conversational goals.

Having identified the model with the highest agreement with human judgements in zero-shot settings, we then explored various few-shot prompting strategies to further enhance alignment. We compared two example selection methods: first-$k$ selection, which uses the initial dialogue turns as exemplars; and random sampling, which draws examples from the entire conversation. The effectiveness of each model and prompting strategy was measured by calculating the correlation between predicted ratings and the ground-truth human ratings from Experiment 1.

\paragraph{Evaluation on Additional Scenarios}
We used the LLM and prompting strategy that achieved the highest agreement with human judgements to rate responses in three additional scenarios. We fitted a single OLS regression model to the ratings from all three scenarios. The model examined the effects of the response condition (LLM-generated vs. human-authored), dialogue turn, and specific scenario, as well as their interactions, on the predicted quality scores. We plotted the individual data points and OLS trend lines for each scenario separately to visualise the underlying patterns within each context.

\subsection{Task 2: Pairwise Preference}

\paragraph{Validation on Ground-Truth}
In parallel, we validated the use of an \textit{LLM-as-a-judge} for a pairwise preference task. We used the \mbox{Gemini 2.0 Flash} model, providing it with complete contextual information and, at each turn, both the human-authored and LLM-generated response. The model chose the response that best fit the conversation.

To simulate the variance found in human judgements, we set the model’s \textit{temperature} hyperparameter to 1.2 to increase response stochasticity. We evaluated each response pair 50 times, allowing us to compute a preference proportion for each option. To prevent order bias, the presentation order of the two responses was randomised in each prompt. The reliability of this approach was validated by correlating the LLM preference proportions with the human preference proportions from Experiment 1.

\paragraph{Evaluation on Additional Scenarios}
\label{subsubsec:exp2_generalization}

Given the strong prior alignment with human preferences using \textsc{Gemini 2.0 Flash} in validation, we applied zero-shot prompting for pairwise preference judgements in the additional scenarios. We prompted the model to indicate its preference for each response pair, and computed the proportion of preferences for the LLM-generated responses and plotted the linear trend using a simple OLS model.

\subsection{Results}
\subsubsection{Task 1: Construct Ratings}
\paragraph{Validation}
We evaluated the reliability of using \textit{LLM-as-a-judge} across various LLMs with varied capabilities and prompting strategies. Table~\ref{tab:correlation_results} shows the Pearson ($r_p$) and Spearman ($r_s$) correlation coefficients for each distinct setup. 

In a zero-shot setting, more advanced models serve as more reliable evaluators. For instance, \textsc{Gemini 2.0 flash} achieves the highest alignment on \textit{Overall Quality} ($r_p=0.519, r_s = 0.452$), substantially outperforming smaller models like \textsc{Llama 3.1} ($r_p=0.232$). Besides, less capable models appear to be correlating positively with human judgements on LLM-generated responses but negatively on human-authored text (e.g., \textsc{Llama 3.1} on \textit{Natural}: $r_p=0.302$ vs. $-0.178$). In contrast, the most capable model, \textsc{Gemini 2.0 flash}, demonstrated strong alignment on human-authored responses with the highest correlation on \textit{Overall Quality} ($r_p=0.632$) but presented no correlation on LLM-generated responses ($r_p=-0.001$). 

Building upon the best-performing \textsc{Gemini 2.0 flash} model, we further investigate how prompting strategies affect alignment with human judgements. Notably, no correlation is reported for \textit{Understandable} among all strategies due to a lack of variance in ratings. The LLM consistently assigned the maximum score, while human judges also gave high scores with minimal variation. This ceiling effect suggests that responses were perceived as highly understandable.

Adopting a few-shot prompting strategy significantly enhances \textsc{Gemini 2.0 flash}'s alignment with human judgements. Random Sampling 6-shot was the most effective strategy ($r_p = 0.659, r_s = 0.682 $ on \textit{Overall Quality}), slightly outperforming the prompting strategy using the first six turns as examples ($r_p = 0.629, r_s = 0.650$). Besides, the increased sample size and introduced randomness in example selection contribute to improved alignment with human judgements on LLM-generated responses, with $r_p$ gradually increasing from -0.001 to 0.659 and $r_s$ increasing from 0.119 to 0.632.

\paragraph{Evaluation}
As the random sampling 6-shot prompting showed relatively stronger correlations across all response conditions ($r_p = 0.659, r_s = 0.682)$, we applied this strategy to obtain ratings for three additional scenarios. 

An OLS regression was conducted to analyse the construct ratings across the three additional scenarios. The overall model was statistically significant, explaining approximately 21.8\% of the variance in the scores ($F(11, 202) = 5.13, p < .001, R^2 = .218$). The analysis revealed a significant positive main effect for \texttt{turn} specifically for the baseline human-authored condition ($\beta = 0.019, p = .003$), indicating increasing perceived quality in human-authored responses as the dialogues progressed. 

While there was no significant main effect for the response \texttt{condition}, suggesting comparable average performance, we observed a marginally significant interaction between \texttt{condition} and \texttt{turn} ($\beta = -0.017, p = .067$). This negative interaction, visualised in Figure \ref{fig:cons_plots}, indicated that the quality trend for LLM-generated responses was significantly less positive than for human-authored responses over time. No other interactions involving the different scenarios were significant, suggesting this pattern was consistent across additional contexts. 

\subsubsection{Task 2: Pairwise Preference}
\paragraph{Validation}
To examine the reliability of the \textit{LLM-as-a-judge} for the pairwise preference task, we correlated the LLM's preference distributions with the human preference proportions. The relationship was found to be significantly positive for both Pearson’s correlation ($r = .656, p < .001$) and Spearman’s correlation ($r_s = .643, p = .003$), which suggests high agreement with human judgements on pairwise comparisons.

\paragraph{Evaluation}
The extensive evaluation revealed that the \textit{LLM-as-a-judge} held a consistent and strong preference for human-authored responses across all additional scenarios. This is visually demonstrated in Figure \ref{fig:pre_plots}, where the majority of conversational turns show a preference proportion below the 50\% threshold for the LLM-generated option. While a clear preference level was established, a subsequent regression analysis found no statistically significant trend in these preferences over the course of the dialogues.
\section{Discussion and Conclusions}
Answering our research questions led to three principal findings. First, we observed significant degradation in the perceived quality of LLM-generated responses relative to human-authored responses as knowledge-grounded role-play dialogues progress. This trend was validated by both human evaluation and our automated \textit{LLM-as-a-judge} assessments and was most prominent in the human-rated dimensions of \textit{Overall Quality}, \textit{Natural}, and \textit{Context Maintenance}.

Second, multi-turn analysis revealed opposing trajectories and confirmed that LLM performance degrades sharply over turns, while human-authored responses improved throughout the conversation (\textbf{SubRQ1}). This aligns with prior work on context degradation in extended interactions \cite{liu_lost_2024} but expands it to knowledge-grounded pedagogical settings. Third, participants’ quality perceptions were most strongly driven by \textit{Interesting}, \textit{Maintains Context}, and \textit{Natural} (\textbf{SubRQ2}). Experts in our focus group highlighted additional nuances: character consistency, pedagogical nudging and conciseness. These factors resonate with the design principles of educational role-play systems \cite{gousseva_creating_2024} but have not been quantitatively assessed.

To scale our investigation, we validated the \textit{LLM-as-a-judge} approach \cite{zheng_judging_2023} for both construct rating and pairwise preference tasks. Using \textsc{Gemini 2.0 Flash} with a random few-shot prompting strategy, we achieved high alignment with human judgements, providing a viable method for automated evaluation of long-form dialogues. The consistent preference for human-authored responses in pairwise evaluations further reinforces the reliability of LLM judges for comparative assessments, as suggested by \citet{park_paireval_2024}. This evaluation framework also provides a methodology for benchmarking specialised judging models, such as Prometheus-2 \cite{kim_prometheus_2024} and JudgeLM \cite{zhu_judgelm_2025}. Notably, however, automated evaluation revealed that LLMs struggle to evaluate output of their kind, as seen in the lower correlations for LLM-generated responses in zero-shot settings. This calls for caution when choosing LLM judges for evaluating LLM performance. 

Overall, our findings highlight both the potential and current limitations of LLMs in training-oriented role-play dialogues. Despite our validated evaluation pipeline that enables scalable assessment, LLMs still struggle to sustain high-quality, context-sensitive responses across extended interactions. This long-context degradation remains a significant barrier. Until such challenges are addressed, human authors continue to be the gold standard for crafting engaging and pedagogical role-play scenarios.

\section*{Limitations}
This study has several limitations. First, human evaluation was conducted on a single scenario. Although we expanded to additional scenarios in the automated evaluation, the generalisability of our human findings to other domains remains to be tested. 

Second, our \textit{LLM-as-a-judge} approach, while effective, exhibited limitations. The \textit{Understandable} dimension showed a ceiling effect in both human and automated evaluations, limiting its discriminative power.
In zero-shot settings, the LLM judge (\textsc{Gemini 2.0 Flash}) showed lower alignment with human judgements for LLM-generated responses than for human-authored ones, suggesting potential biases. Few-shot prompting mitigated this, but the approach requires careful calibration of example selection and may not transfer seamlessly to other models or tasks.

Third, our study focused on a specific LLM (\textsc{Llama 3}) for response generation. This choice enabled controlled comparisons and the ability to fine-tune on high-quality training scenarios, which is not feasible with more powerful proprietary models like \textsc{Gemini 2.0} or \textsc{Claude 3}. However, this focus does not capture potential differences arising from variations in model architecture, scale, or fine-tuning approaches. While stronger models potentially handle long-context modelling better, mitigating some of the degradation effects we observed, not being able to fine-tune these models constrains their adaptability for pedagogical customisations on training scenarios.

These limitations suggest future research directions, including investigating the applicability of our findings to a wider variety of dialogue scenarios, improving the robustness of using \textit{LLM-as-a-judge} and exploring potential strategies to mitigate performance decay over extended interactions. Beyond addressing these limitations, a promising future direction is to broaden the analytical approach, examining the dialogue sub-structures for more nuanced, qualitative insights into the differences between human and LLM-generated responses.

\section*{Ethics Statement}

This study complies with the research ethics guidelines of Utrecht University. The research protocol was assessed through the university's standard ethics screening procedure (the \textit{Ethics and Privacy Quick Scan}) and classified this research as low-risk with no further ethics review or privacy assessment required.

The study's sample included participants recruited via Prolific and a convenience sample from personal contacts. All participants provided informed consent, detailing the task, data collection, and their right to withdraw without penalty. Prolific participants were compensated at £9.20/hr, meeting platform guidelines. The convenience sample participants received non-monetary gratuities. The research used non-sensitive fictional negotiation dialogues with no foreseeable risks. All personally identifiable information was anonymised. 

\section*{Acknowledgements}
This activity is partly funded by the PPP subsidy from the Dutch Ministry of Economic Affairs and Climate Policy through CLICKNL. CLICKNL is the leading consortium for Knowledge and Innovation (TKI) in the Creative Industries in the Netherlands.

\bibliography{bibliography}

\newpage
\appendix
\clearpage
\onecolumn

\appendix
\section{Model Details}
This section outlines the language models used in our experiments, including their deployment, source checkpoints, key specifications, and the fine-tuning hyperparameters for the response generation model.

\begin{table}[H]
\centering
\begin{threeparttable}
    \caption{Language models and their configurations used in the experiments.}
    \label{tab:model_comparison}
    \renewcommand{\arraystretch}{1.2}
    \small
    \begin{tabularx}{1.00\textwidth}{l l X c}
        \toprule
        \textbf{Model} & \textbf{Deployment} & \textbf{Checkpoint} & \textbf{Params} \\
        \midrule
        \textsc{Llama 3}\tnote{a} & Hugging Face (Fine-tuned) & \texttt{unsloth/llama-3-8b-bnb-4bit} & 8.03B \\
        \textsc{Llama 3.1}\tnote{b} & Hugging Face & \texttt{meta-llama/Meta-Llama-3.1-8B-Instruct} & 8.03B \\
        \textsc{Mistral}\tnote{b} & Hugging Face & \texttt{mistralai/Mistral-7B-Instruct-v0.2} & 7.24B \\
        \textsc{Phi-3 Medium}\tnote{b} & Ollama & \texttt{phi3:14b} & 14.00B \\
        \textsc{Gemini 2.0 Flash}\tnote{c} & Vertex AI API & (Proprietary) & N/A \\
        \bottomrule
    \end{tabularx}

    \begin{tablenotes}[para]
        \small
        \item[a] \textsc{Llama 3} was fine-tuned on our dataset for response generation, see Table \ref{tab:hyperparameters} for details. The other models listed served as judges to evaluate the generation output.
        \item[b] The judging models (\textsc{Llama 3.1}, \textsc{Mistral}, and \textsc{Phi-3 Medium}) were used in a zero-shot setting.
        \item[c] \textsc{Gemini 2.0 Flash} was used in both zero-shot and few-shot settings for comparison.
    \end{tablenotes}
\end{threeparttable}
\end{table}

\begin{table}[h!]
\centering
\caption{Hyperparameters for fine-tuning the \textsc{Llama 3} model.}
\label{tab:hyperparameters} 
\begin{tabular}{ll}
\hline
\textbf{Parameter} & \textbf{Value} \\
\hline
\multicolumn{2}{l}{\textit{\textbf{Fine-tuning Parameters}}} \\
LoRA ($r$, $\alpha$) & (16, 16) \\
Optimiser & AdamW (8-bit) \\
Learning Rate & $2 \times 10^{-4}$ \\
Batch Size & 4 \\
Epochs & 9 \\
\midrule
\multicolumn{2}{l}{\textit{\textbf{Inference Parameters}}} \\
Temperature & 1.0 \\
Top-p ($p$) & 0.3 \\
Top-k ($k$) & 10 \\
\hline
\end{tabular}
\end{table}

\newpage
\section{Generation Prompt}
\label{sec:generationprompt}
This section describes the prompt used for response generation. At each turn, the model receives contextual information, including the scenario background, dialogue history, and character role descriptions, to guide its output.

\begin{figure}[H]
    \centering
    \begin{minipage}{\textwidth}
    \begin{lstlisting}
Your function is to assist in the creation and completion of a Conversational Scenario. In every Conversational Scenario, there are playerStatements, computerStatements. Every Conversational Scenario has:

- Player's Role: defines the role of the player when playing the scenario.
- Character's Role: defines the personality and objectives of the digital persona interacting with the player.

Your role specifically involves generating computerStatements sequences that could respond to the last playerStatement. Your computerStatements sequences must respond to the last playerStatement, but also make sense when read before the next playerStatement.

After putting all together, the conversational Scenario should be readable and understandable with a nice transition between statements. Some important rules for the generation of computerStatements are:

- computerStatements are generated in sequences and should be readable one after the other.
- The computerStatements should be as short as possible, and they shouldn't be longer than one sentence.

---- Main Task ---
Help me complete the following conversational Scenario by generating the computerStatements sequences between the last playerStatement and the next playerStatement.

---- Role Description ----
Player's Role:
Character's Role:

---- Dialogue Context ----
Previous statements:
Last playerStatement:
Next playerStatement`:
\end{lstlisting}
    \end{minipage}
    \caption{The prompt template for response generation.}
    \label{fig:prompt_generation}
\end{figure}

\newpage
\section{Example Dialogue Comparison}

The following table presents human-authored and LLM-generated responses in 23 agent turns from the negotiation scenario.

\begin{longtable}{c p{7cm} p{7cm}}

    % --- Caption and Header Definition ---
    
    \caption{Comparisons of Human-Authored and LLM-Generated agent responses from the negotiation scenario.}
    \label{tab:dialogue_example} \\
    \toprule
    \textbf{Turn} & 
    \multicolumn{1}{c}{\textbf{Human-Authored}} & 
    \multicolumn{1}{c}{\textbf{LLM-Generated}} \\ 
    \midrule
    \endfirsthead % Marks the end of the header for the VERY FIRST page
    
    \toprule
    \multicolumn{3}{c}{\tablename~\thetable: Continued from previous page} \\ % Optional: Adds a "continued" notice
    \midrule
    \textbf{Turn} & 
    \multicolumn{1}{c}{\textbf{Human-Authored}} & 
    \multicolumn{1}{c}{\textbf{LLM-Generated}} \\ 
    \midrule
    \endhead % Marks the end of the header for ALL SUBSEQUENT pages
    
    \bottomrule
    \endlastfoot % Marks the footer for the VERY LAST page

        & \multicolumn{2}{p{14cm}}{Agent: Hi, it's good to see you! I look forward to advancing our wonderful collaboration.} \\
        & \multicolumn{2}{p{14cm}}{User: Yes, same here. I look forward to exploring solutions with you.} \\
        \midrule
        
        1 & Agent: Me too. So how are you? & Agent: Me too. In fact, I do believe our cooperation is a good reason to be optimistic. \\
        \midrule
        
        ~ & \multicolumn{2}{p{14cm}}{User: I'm good, and happy to see you guys eventually solved the problems with your update.} \\
        \midrule
        
        2 & Agent: Yes, I'm glad we managed to solve that. & Agent: Yes, we were able to resolve the issue. \\
        \midrule
        
        & \multicolumn{2}{p{14cm}}{User: And you successfully managed to prevent a data leak, which is important.} \\
        \midrule
        
        3 & Agent: We're content with how that went. As it should naturally. & Agent: Yes, of course. As a company, we take that very seriously. \\
        \midrule
        
        & \multicolumn{2}{p{14cm}}{User: The problems raised some concerns about the prolongation within our organisation, but personally, I think you did well in recovering.} \\
        \midrule
        
        4 & Agent: - Well, thank you. - And I do think we recovered well. Business is booming. - So let's see what we can do for you. & Agent: That's good to hear. We strive for continuous improvement. \\
        \midrule
        
        & \multicolumn{2}{p{14cm}}{User: I'd like to explore opportunities to extend our contract, in the context of our future relationship.} \\\midrule

        5 & Agent: - Alright then. - But I hope you agree that our solution continues to be a perfect match for your organisation. & Agent: Sure, that's fine, but that was already agreed. \\
        \midrule

        & \multicolumn{2}{p{14cm}}{User: Your software indeed matches some of our needs, but it's not on par with several other providers.} \\\midrule

        6 & Agent: - Oh, okay. - I can’t imagine you are referring to all the new features. - I’m sure your team will be quite happy with what we did. & Agent: - Oh, okay. - I would like to ask you to be more specific about what you mean.\\
        \midrule

        & \multicolumn{2}{p{14cm}}{User: Can you tell me more about your reasons for not supporting functionalities for our invoicing process?} \\\midrule

        7 & Agent: - We still intend to include it in the future. - We postponed this for now because you were the only client who asked for this. I could consult the team about including it in the next version. & Agent: Well, that's a good question. I'll have to check with our developers.\\
        \midrule

        & \multicolumn{2}{p{14cm}}{User: It's becoming a market standard and we now have to solve it ourselves.} \\\midrule

        8 & Agent: Yes, but in those packages, invoicing facilities are not perfect and by no means a guarantee that it will work in your environment. & Agent: - I see, but we can't do much about that now. - If you want this feature, maybe we should consider taking it from a different provider.\\
        \midrule

        & \multicolumn{2}{p{14cm}}{User: I might not disagree with you, but it is a disappointment that it's not included in this version.} \\\midrule

        9 & Agent:  I get that, but I still think our software is quite good for your needs. & Agent: - I would have hoped you would see the value of the software we provide. - If you want more, you will have to take this into account.\\
        \midrule

        & \multicolumn{2}{p{14cm}}{User: The cost to connect our invoicing is about €300k, which I do need to find somewhere.} \\
        \midrule

        10 & Agent: - There isn't much we can do. - But I guarantee excellent support for development on your side. & Agent: That's still a lot of money, but we do this for all our large clients.\\
        \midrule

        & \multicolumn{2}{p{14cm}}{User: How would you guarantee this?} \\\midrule

        11 & Agent: - I see your point. - Our support team will help contribute to this development. - So we'll include this in the contract. & Agent: Well, that's what we do with all our customers.\\
        \midrule

        & \multicolumn{2}{p{14cm}}{User: Naturally, but it still means lots of work for us, which does need compensation.} \\\midrule

        12 & Agent: - Fair enough, but you have to understand we are already stretched with our offer. - I might be able to provide a consultant free of charge for two weeks, but that's a big stretch. & Agent: You have to see it my way, too. We can't give discounts.\\
        \midrule

        & \multicolumn{2}{p{14cm}}{User: I appreciate the suggestion, but it still requires compensation.} \\\midrule

        13 & Agent: What I can do is find a substitute for this functionality in Excel. If this is indeed favourable to you. & Agent: I'll see what I can do, but it's not a decision I can make.\\
        \midrule

        & \multicolumn{2}{p{14cm}}{User: Can you elaborate on how your offer would be favorable to us?} \\\midrule

        14 & Agent: We don't compromise on service. That's also what you want, isn't it? & Agent: I'll see what I can do, but it's not something I promise.\\
        \midrule

        & \multicolumn{2}{p{14cm}}{User: Most suppliers provide good service, so let's discuss the price increase.} \\\midrule

        15 & Agent: - Yes, well, okay, let's look at that. -  Like I said in my email, our prices have increased by 24\%. But for you, I have managed to limit this to 18\%. - This increase is reasonable and it is legitimate. & Agent: - You don't think I do what I do for free, do you? We have to keep covering our costs. - You won't find a supplier that meets our prices.\\
        \midrule

        & \multicolumn{2}{p{14cm}}{User: My understanding of the market development is not in that range.} \\\midrule

        16 & Agent: As I wrote in our quote, the 18\% includes compensation for 8\% inflation last year, combined with all the new functionalities. & Agent: So, what would be your proposal?\\
        \midrule

        & \multicolumn{2}{p{14cm}}{User: Unfortunately, the new functionalities don't fulfil all our needs.} \\\midrule

        17 & Agent: - Maybe not yet. - Nevertheless, I am confident you'll find them useful, especially our new automation of the approval process. & Agent: - Well, I can't do anything about that. - If you don't see the value, maybe we should consider our proposal as a chance to get a discount.\\
        \midrule

        & \multicolumn{2}{p{14cm}}{User: I'm navigating by what the business tells me, that these functions have limited value.} \\\midrule

        18 & Agent: - I think they will see the value of the new functionalities when they start using them. - Okay. - We could perhaps do 16\%, but that is really pushing it. & Agent: - Well, I would really be sorry to see you go. You are our most important client. - I'm sure we can find a compromise.\\
        \midrule

        & \multicolumn{2}{p{14cm}}{User: The need for the Excel workaround impacts our budget significantly, which means I can't go beyond 13\%.} \\\midrule

        19 & Agent: I do see your point here, but I can't go lower than 16\% at the moment. & Agent: - I can't go below 15\%. - I'll see what I can do, but it's difficult to make this happen.\\
        \midrule

        & \multicolumn{2}{p{14cm}}{User: We're contemplating extending for one year, but what if we consider prolonging for two years for 12\%?} \\\midrule

        20 & Agent: - That indeed would give us some more room to manoeuvre. Naturally, we'd be happy to work something out towards two years. - But I doubt we could go below 14\%.& Agent: - That would be more like it. - On a two-year contract with the same conditions, I could probably give you a 14\% discount.\\
        \midrule

        & \multicolumn{2}{p{14cm}}{User: How do you want me to sell this price increase with a two-year extension to my stakeholders?} \\\midrule

        21 & Agent: - I would need to follow this up internally. - But I'll see what I can do towards 14\% based on a two-year prolongation. & Agent: Well, you have a good point there. Shall we agree at 14\%?\\
        \midrule

        & \multicolumn{2}{p{14cm}}{User: Okay, so you'll commit to 14\% for two years.} \\\midrule

        22 & Agent: Yes, if the extension is two years, I think I can commit to that. & Agent: Yes, if that is what it needs to be, then of course.\\
        \midrule

        & \multicolumn{2}{p{14cm}}{User: If you can send me your updated offer, including the support and 14\% increase, then I will share this with my stakeholder.} \\\midrule

        23 & Agent: - Of course. - I'm glad we've managed to figure this out. So let's put this in writing. We look forward to working with you for the coming two years. & Agent: That sounds good, I'll see to it that it is sent today.\\
        \bottomrule
\end{longtable}

\newpage
\section{In-simulation Interaction}
This section presents screenshots of in-simulation interactions where participants were shown two responses, one LLM-generated and one human-authored, within a given context and asked to indicate their preference at each turn.

\begin{figure}[H]
    \centering
    \includegraphics[width=0.7\linewidth]{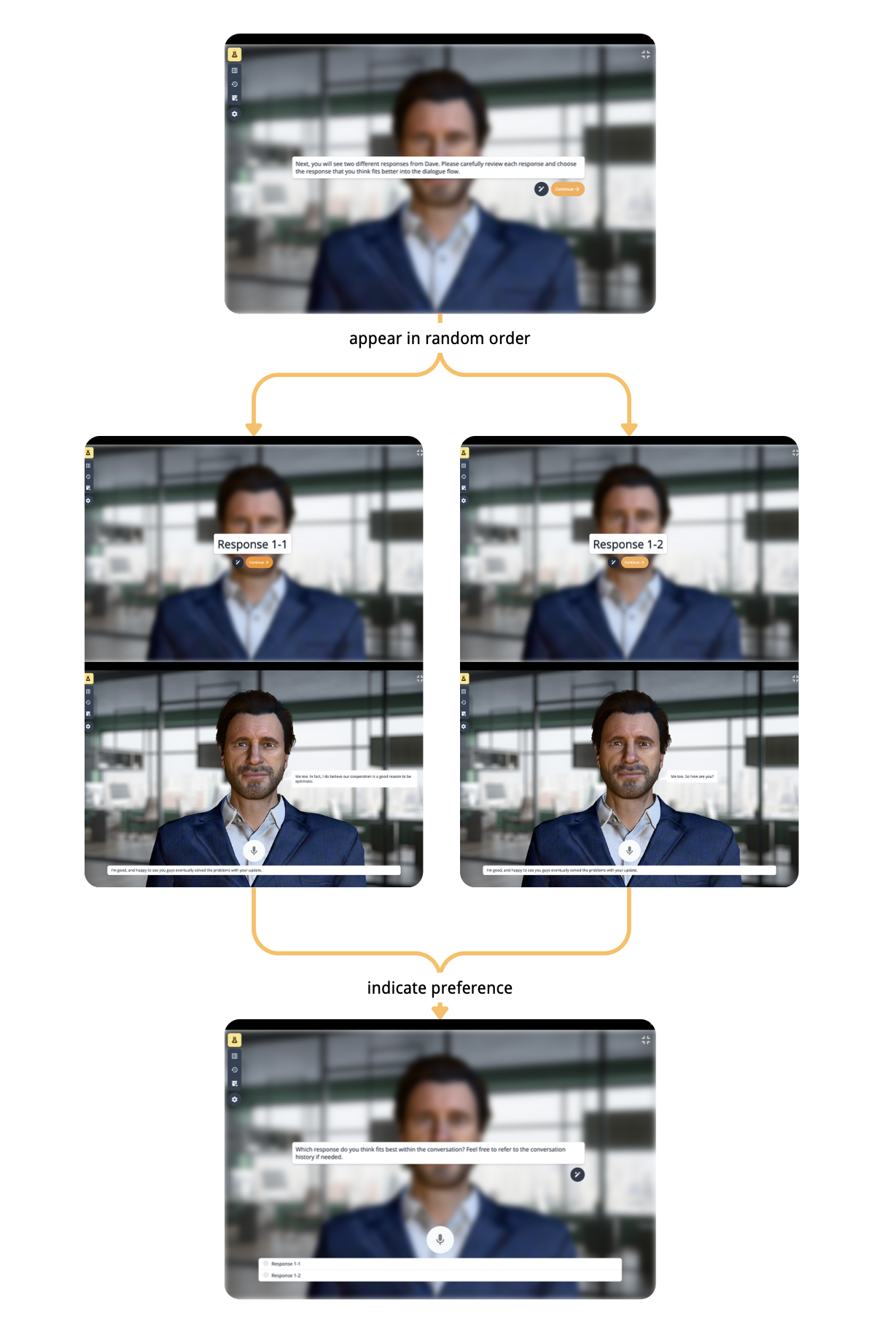}
    \caption{In-simulation pairwise preference indication.}
    \label{fig:pre_procedure}
\end{figure}

\newpage
\section{Human Evaluation Results}
This section presents the complete statistical results of the LMM analysis for human-rated quality constructs.

\begin{table*}[h!]
\centering
\caption{Full Results of the LMMs for each quality construct. The table displays the fixed-effect coefficients ($\beta$), standard errors ($SE$), $z$-values, and $p$-values for the main effect of \texttt{Condition} (LLM-generated vs. Human-authored) and the \texttt{Condition × Turn} interaction.}
\label{tab:lmm_results}
\begin{tabular}{llcccc}
\hline
\textbf{Construct} & \textbf{Effect} & \textbf{$\beta$} & \textbf{$SE$} & \textbf{z-value} & \textbf{p-value} \\ \hline
\textit{Understandable} & \texttt{Condition} & 0.034& 0.032 & 1.057 & .291\\
                        & \texttt{Turn} & 0.004  & 0.002  & 2.545  & .011 \\
                        & \texttt{Condition × Turn} & -0.005 & 0.002 & -2.022 & .043\\ \hline
\textit{Natural}        & \texttt{Condition} & 0.102 & 0.074 & 1.377 & .169\\
                        & \texttt{Turn} & 0.016  & 0.004  & 4.267  & <.001 \\
                        & \texttt{Condition × Turn} & -0.021 & 0.005 & -3.863 & <.001\\ \hline
\textit{Maintains Context} & \texttt{Condition} & 0.095 & 0.074 & 1.292 & .196\\
                        & \texttt{Turn} & 0.016  & 0.004  & 4.135  & <.001 \\
                        & \texttt{Condition × Turn} & -0.020 & 0.005 & -3.706 & <.001\\ \hline
\textit{Interesting}    & \texttt{Condition} & 0.007 & 0.083 & 0.083 & .934\\
                        & \texttt{Turn} & 0.022  & 0.004  & 5.250  & <.001 \\
                        & \texttt{Condition × Turn} & -0.018 & 0.006 & -2.930 & .003\\ \hline
\textit{Uses Knowledge} & \texttt{Condition} & 0.114 & 0.041 & 2.772 & .006\\
                        & \texttt{Turn} & 0.012  & 0.002  & 5.449  & <.001 \\
                        & \texttt{Condition × Turn} & -0.013 & 0.003 & -4.371 & <.001\\ \hline
\textit{Overall Quality} & \texttt{Condition} & 0.008 & 0.117 & 0.069 & .945\\
                        & \texttt{Turn} & 0.027  & 0.006  & 4.446  & <.001 \\
                        & \texttt{Condition × Turn} & -0.029 & 0.009 & -3.363 & .001\\ \hline
\end{tabular}
\end{table*}

\newpage
\section{Automated Evaluation Results}
This section supplements human evaluation with \textit{LLM-as-a-judge} analysis across two tasks on additional scenarios: construct ratings and pairwise preference.

\subsection{Task 1: Construct Ratings}
This section reports Pearson ($r_p$) and Spearman ($r_s$) correlation coefficients for construct ratings across LLM models, prompting strategies, and response conditions (see Appendix \ref{subsubsec:validation}). We then visualise LLM-predicted ratings on additional scenarios using the configuration that showed the highest agreement with human judgments in Appendix \ref{subsubsec:evaluation}.

\subsubsection{Validation}
\label{subsubsec:validation}
\begin{table}[H]
\centering
\caption{Pearson ($r_p$) and Spearman ($r_s$) Correlation Coefficients by Prompting Strategy, Construct and Response Condition}
\label{tab:correlation_results}
\resizebox{\textwidth}{!}{%
\begin{tabular}{@{}llcccccccccccc@{}}
\toprule
& & \multicolumn{2}{c}{\textbf{Understandable}} & \multicolumn{2}{c}{\textbf{Natural}} & \multicolumn{2}{c}{\textbf{Maintains Context}} & \multicolumn{2}{c}{\textbf{Interesting}} & \multicolumn{2}{c}{\textbf{Uses Knowledge}} & \multicolumn{2}{c}{\textbf{Overall Quality}} \\
\cmidrule(lr){3-4} \cmidrule(lr){5-6} \cmidrule(lr){7-8} \cmidrule(lr){9-10} \cmidrule(lr){11-12} \cmidrule(lr){13-14}
\textbf{Configuration} & \textbf{Response Condition} & \textbf{$r_p$} & \textbf{$r_s$} & \textbf{$r_p$} & \textbf{$r_s$} & \textbf{$r_p$} & \textbf{$r_s$} & \textbf{$r_p$} & \textbf{$r_s$} & \textbf{$r_p$} & \textbf{$r_s$} & \textbf{$r_p$} & \textbf{$r_s$} \\
\midrule
\multirow{3}{*}{Zero-shot (\textsc{Mistral-7B})} & General & 0.098 & 0.082 & 0.230 & 0.165 & 0.398 & 0.367 & 0.308 & 0.343 & 0.024 & 0.055 & 0.308 & 0.230 \\
& LLM-generated & 0.273 & 0.276 & 0.345 & 0.259 & 0.441 & 0.437 & 0.221 & 0.260 & 0.091 & 0.130 & 0.333 & 0.188 \\
& Human-authored & -0.164 & -0.208 & 0.144 & 0.195 & 0.364 & 0.339 & 0.248 & 0.284 & -0.081 & -0.024 & 0.204 & 0.260 \\

\multirow{3}{*}{Zero-shot (\textsc{Llama 3.1 8B})} & General & -0.053 & -0.163 & 0.122 & -0.113 & -0.012 & 0.034 & 0.111 & 0.179 & 0.088 & 0.119 & 0.232 & 0.099 \\
& LLM-generated & 0.106 & 0.078 & 0.302 & 0.020 & 0.059 & 0.101 & 0.292 & 0.347 & 0.160 & 0.217 & 0.168 & 0.016 \\
& Human-authored & -0.411 & -0.426 & -0.178 & -0.256 & -0.207 & -0.065 & 0.159 & 0.130 & -0.076 & -0.044 & 0.175 & 0.162 \\

\multirow{3}{*}{Zero-shot (\textsc{Phi-3 Medium 14B})} & General & 0.120 & 0.012 & 0.227 & 0.181 & 0.105 & 0.059 & 0.086 & 0.017 & 0.461 & 0.430 & 0.270 & 0.254 \\
& LLM-generated & 0.348 & 0.299 & 0.166 & 0.094 & 0.118 & 0.089 & -0.204 & -0.191 & 0.344 & 0.400 & 0.147 & 0.086 \\
& Human-authored & -0.266 & -0.278 & 0.202 & 0.183 & 0.077 & 0.074 & 0.157 & 0.144 & 0.512 & 0.370 & 0.230 & 0.227 \\

\multirow{3}{*}{Zero-shot (\textsc{Gemini 2.0 Flash})} & General & nan & nan & 0.157 & 0.166 & -0.279 & -0.063 & 0.298 & 0.345 & 0.478 & 0.520 & 0.519 & 0.452 \\
& LLM-generated & nan & nan & 0.096 & 0.072 & -0.247 & -0.162 & 0.012 & 0.155 & 0.126 & 0.312 & -0.001 & 0.119 \\
& Human-authored & nan & nan & 0.178 & 0.266 & -0.256 & -0.048 & 0.265 & 0.322 & 0.840 & 0.594 & 0.632 & 0.555 \\
\midrule

\multirow{3}{*}{First 3-shot (\textsc{Gemini 2.0 Flash})} & General & nan & nan & 0.320 & 0.303 & 0.224 & -0.014 & 0.433 & 0.456 & 0.359 & 0.300 & 0.555 & 0.494 \\
& LLM-generated & nan & nan & -0.076 & 0.076 & 0.227 & -0.277 & 0.229 & 0.342 & 0.347 & 0.187 & 0.144 & 0.220 \\
& Human-authored & nan & nan & 0.396 & 0.327 & -0.101 & -0.260 & 0.277 & 0.270 & 0.127 & 0.234 & 0.538 & 0.489 \\

\multirow{3}{*}{Random Sampling 3-shot (\textsc{Gemini 2.0 Flash})} & General & nan & nan & 0.444 & 0.496 & 0.163 & 0.225 & 0.510 & 0.520 & 0.254 & 0.318 & 0.549 & 0.498 \\
& LLM-generated & nan & nan & 0.081 & 0.254 & -0.229 & -0.252 & 0.213 & 0.166 & 0.206 & 0.262 & 0.281 & 0.259 \\
& Human-authored & nan & nan & 0.592 & 0.566 & 0.391 & 0.370 & 0.425 & 0.560 & 0.283 & 0.353 & 0.562 & 0.506 \\

\multirow{3}{*}{First 6-shot (\textsc{Gemini 2.0 Flash})} & General & nan & nan & 0.295 & 0.402 & 0.301 & 0.316 & 0.531 & 0.564 & 0.272 & 0.441 & 0.629 & 0.650 \\
& LLM-generated & nan & nan & -0.128 & 0.063 & -0.168 & -0.313 & 0.258 & 0.291 & -0.024 & -0.039 & 0.358 & 0.376 \\
& Human-authored & nan & nan & 0.362 & 0.365 & 0.169 & 0.257 & 0.234 & 0.317 & 0.221 & 0.346 & 0.466 & 0.421 \\

\multirow{3}{*}{Random Sampling 6-shot (\textsc{Gemini 2.0 Flash})} & General & nan & nan & 0.406 & 0.442 & 0.233 & 0.252 & 0.482 & 0.576 & 0.211 & 0.418 & 0.659 & 0.682 \\
& LLM-generated & nan & nan & 0.062 & -0.023 & 0.115 & 0.095 & 0.279 & 0.483 & 0.144 & 0.290 & 0.659 & 0.632 \\
& Human-authored & nan & nan & 0.424 & 0.473 & -0.014 & 0.054 & 0.234 & 0.339 & 0.079 & 0.384 & 0.516 & 0.562 \\

\bottomrule
\end{tabular}
}
\end{table}

\newpage
\subsubsection{Evaluation}
\label{subsubsec:evaluation}
\begin{figure}[H]
    \centering 
    \begin{subfigure}[b]{0.6\textwidth} 
        \includegraphics[width=\linewidth]{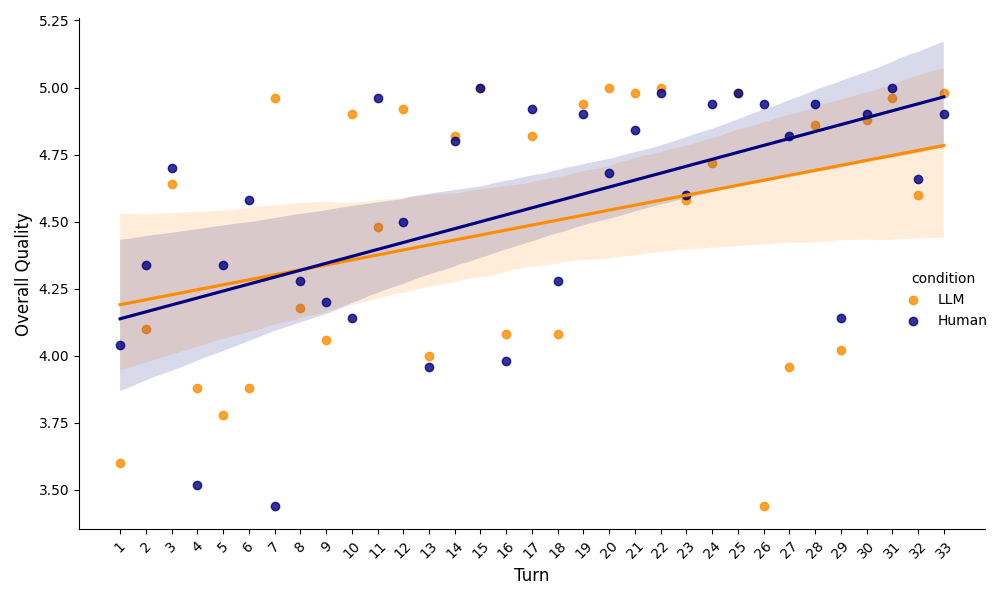}
        \caption{Scenario a: motivational interviewing}
        \label{fig:scenario2}
    \end{subfigure}

    \begin{subfigure}[b]{0.6\textwidth}
        \includegraphics[width=\linewidth]{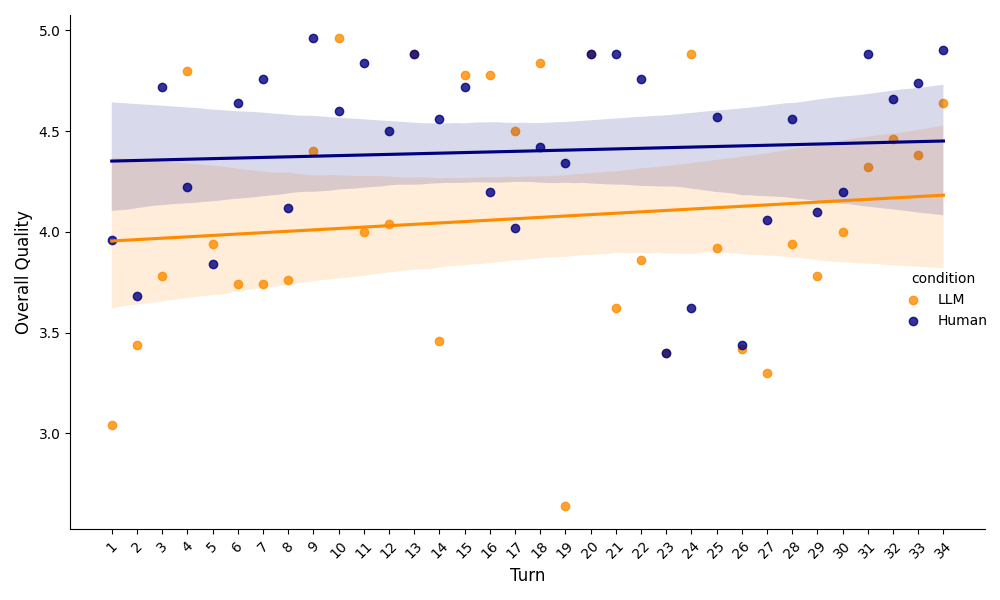}
        \caption{Scenario b: selling}
        \label{fig:scenario3}
    \end{subfigure}

    \begin{subfigure}[b]{0.6\textwidth}
        \includegraphics[width=\linewidth]{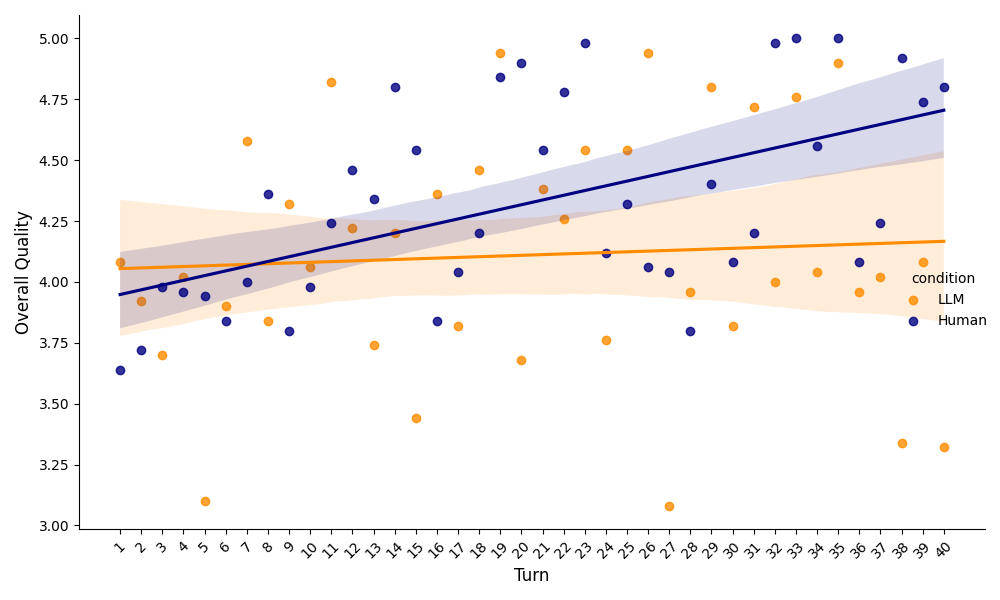}
        \caption{Scenario c: consulting}
        \label{fig:scenario4}
    \end{subfigure}

    \caption{Mean ratings for LLM-generated (orange) and human-authored (blue) responses over time. The diverging trend lines visualise the significant \texttt{Condition × Turn} interaction effect, where the perceived quality of LLM responses degrades relative to the stable performance of human responses.}
    \label{fig:cons_plots}
\end{figure}

\newpage
\subsection{Task 2: Pairwise Preferences}
This section visualises the turn-by-turn preferences of the \textsc{Gemini 2.0 Flash} model on additional scenarios in zero-shot settings.

\begin{figure}[H]
    \centering
    \begin{subfigure}[b]{0.6\textwidth}
        \includegraphics[width=\linewidth]{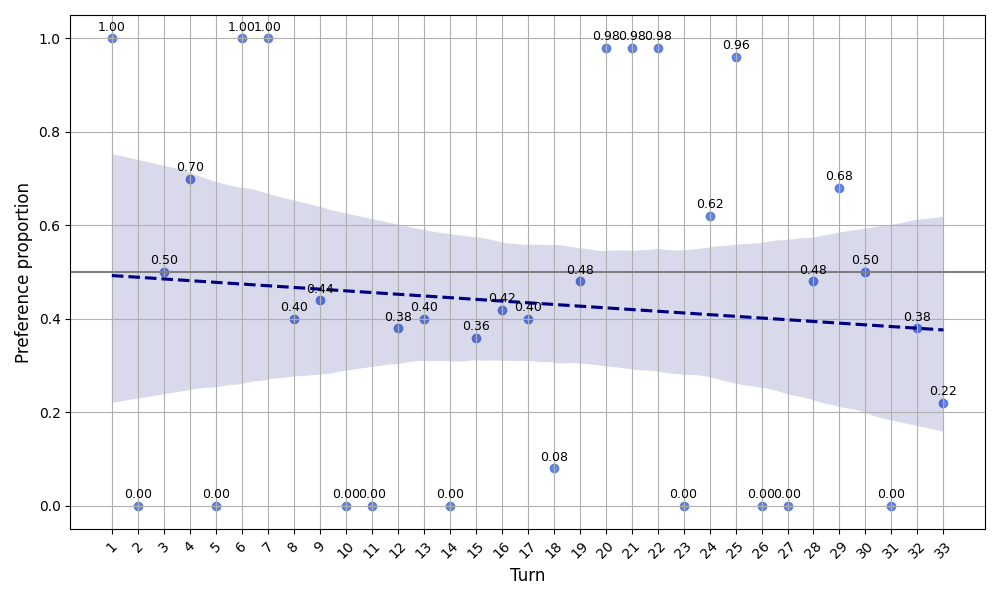}
        \caption{Scenario a: motivational interviewing}
        \label{fig:scenario2}
    \end{subfigure}

    \begin{subfigure}[b]{0.6\textwidth}
        \includegraphics[width=\linewidth]{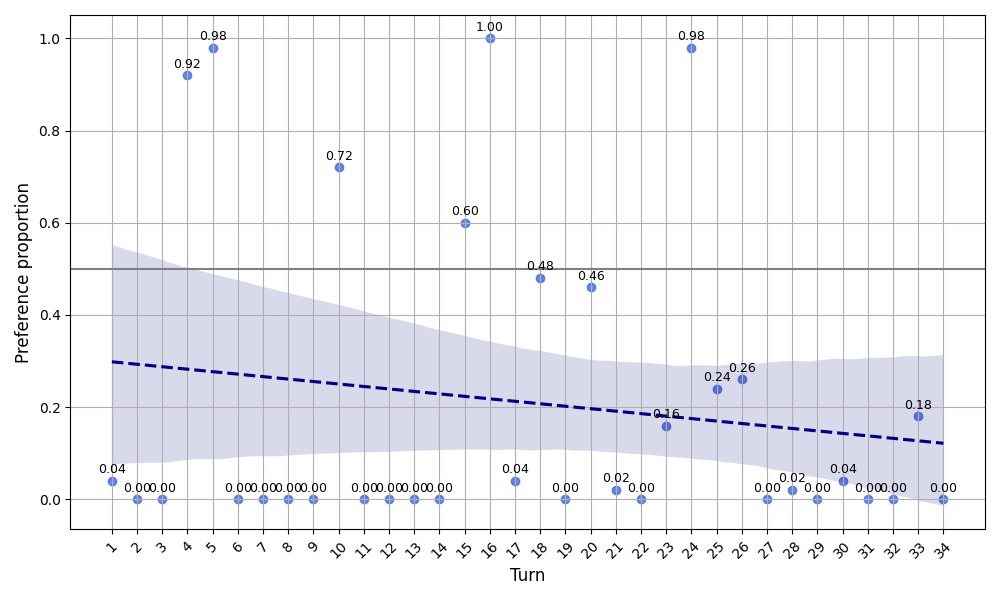}
        \caption{Scenario b: selling}
        \label{fig:scenario3}
    \end{subfigure}

    \begin{subfigure}[b]{0.6\textwidth}
        \includegraphics[width=\linewidth]{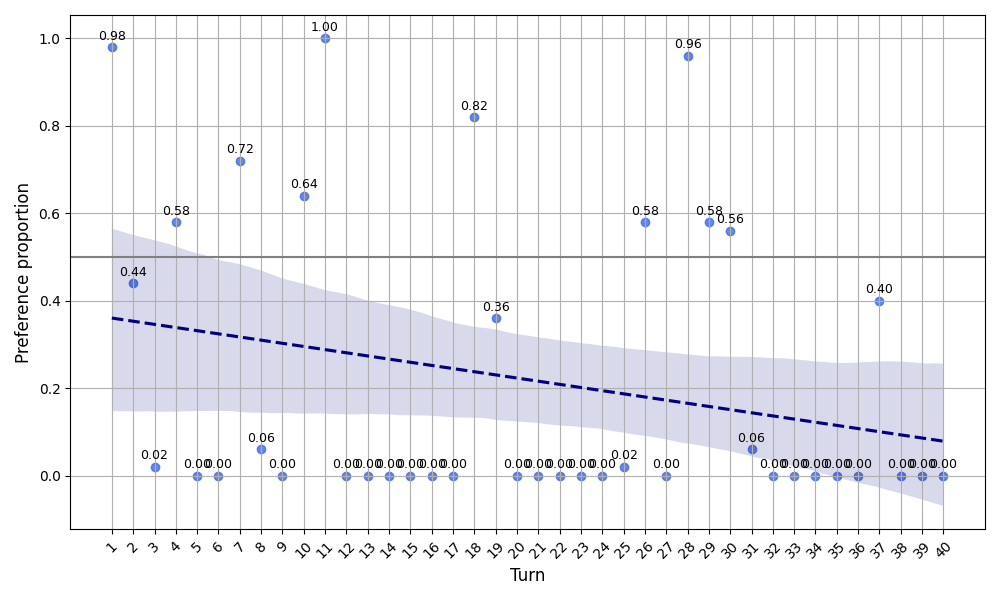}
        \caption{Scenario c: consulting}
        \label{fig:scenario4}
    \end{subfigure}

    \caption{Proportion of preference for the LLM-generated response over turns across additional scenarios. Data points represent the preference proportion at each turn. The OLS trend in dashed line and its shaded 95\% confidence interval illustrate that there was no statistically significant trend over time ($p > .05$), with preferences for the LLM remaining consistently below the 0.5 threshold.}
    \label{fig:pre_plots}
\end{figure}

\end{document}